%% file: arxiv.tex
\documentclass{article}
\usepackage{ijcai22}

\usepackage{times}

\usepackage{soul}
\usepackage{url}
\usepackage{ragged2e}
\usepackage{array}
\usepackage{times}
\usepackage{soul}
\usepackage{dcolumn}
\usepackage{multirow}
\usepackage[hidelinks]{hyperref}
\usepackage[utf8]{inputenc}
\usepackage[small]{caption}
\usepackage{graphicx}
\usepackage{amsmath}
\usepackage{amsfonts}
\usepackage{amssymb}
\usepackage{booktabs}
\usepackage{color}
\usepackage{xcolor}
\usepackage{caption}
\usepackage{subfigure}

\title{WegFormer: Transformers for Weakly Supervised Semantic Segmentation }

\author{
    Chunmeng Liu$^{1}$,
    Enze Xie$^{2}$,
    Wenjia Wang$^{3}$,
    Wenhai Wang$^{4}$, 
    Guangyao Li$^{1}$,
    Ping Luo$^2$ \\
    \affiliations 
    $^1$Tongji University~~~
    $^2$The University of Hong Kong~~~
    $^3$SenseTime~~~~
    $^4$Shanghai AI Lab
    \\
}

\begin{document}

\maketitle

\begin{abstract}

Although convolutional neural networks~(CNNs) have achieved remarkable progress in weakly supervised semantic segmentation~(WSSS), the effective receptive field of CNN is insufficient to capture global context information, leading to sub-optimal results. Inspired by the great success of Transformers in fundamental vision areas, this work for the first time introduces Transformer to build a simple and effective WSSS framework, termed WegFormer. Unlike existing CNN-based methods, WegFormer uses Vision Transformer~(ViT) as a classifier to produce high-quality pseudo segmentation masks. To this end, we introduce three tailored components in our Transformer-based framework, which are (1) a Deep Taylor Decomposition~(DTD) to generate attention maps, (2) a soft erasing module to smooth the attention maps, and (3) an efficient potential object mining~(EPOM) to filter noisy activation in the background. Without any bells and whistles, WegFormer achieves state-of-the-art 70.5\% mIoU on the PASCAL VOC dataset, significantly outperforming the previous best method. We hope WegFormer provides a new perspective to tap the potential of Transformer in weakly supervised semantic segmentation. Code will be released.

\end{abstract}

\section{Introduction}
Semantic segmentation plays an irreplaceable role in many computer vision tasks, such as autonomous driving and remote sensing.
The semantic segmentation community has witnessed continuous improvements in recent years, benefiting from the rapid development of convolutional neural networks (CNNs).
However, expensive pixel-level annotations force researchers to look for cheaper and more efficient annotations to help with semantic segmentation tasks.
Image-level annotations are cheap and readily available, but it is challenging to learn high-quality semantic segmentation models using these weakly supervised annotations.
\emph{This paper focuses on weakly supervised semantic segmentation (WSSS) with image-level annotations.}

Most existing WSSS frameworks~\cite{wei2018revisiting,jiang2019integral} are based on conventional CNN backbones such as ResNet~\cite{he2016deep} and VGG~\cite{simonyan2014very}.
However, the effective receptive field of these CNN-based methods is limited, leading to unsatisfactory segmentation results. 
Recently, Transformer began to show powerful performance in various fundamental areas of computer vision, which was used initially in the natural language process~(NLP). 
Different from CNNs, vision transformers are proven to be able to extract global context information~\cite{xie2021segformer}, which is essential for segmentation tasks and brings fresh thinking to the field of WSSS.  
\begin{figure}[!t]
    \centering
     \includegraphics[scale=0.35]{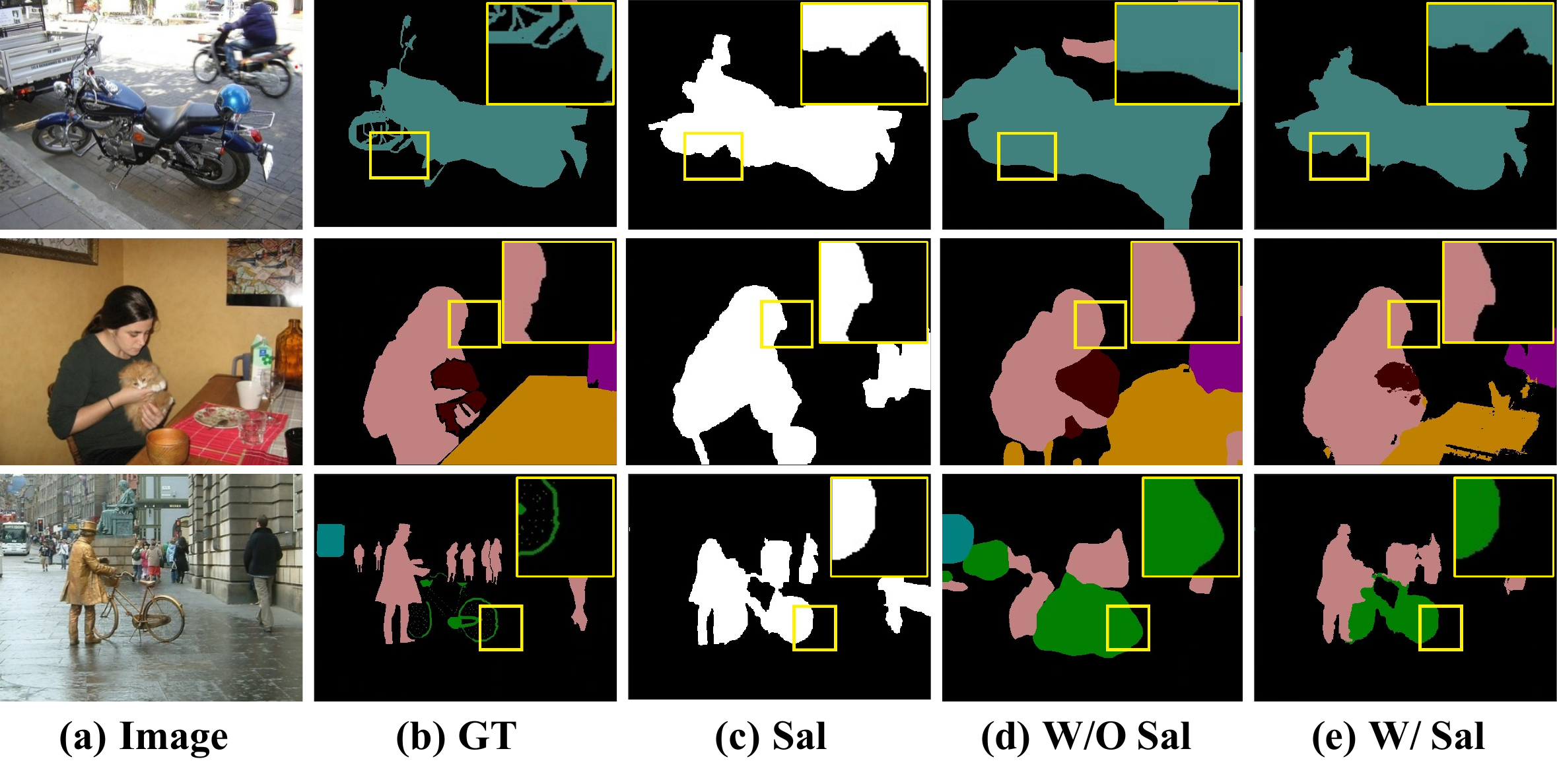}
     \vspace{-3mm}
    \caption{
    Visualization of WegFormer with or without saliency map assist.
    ``Sal'' denotes saliency map.
    From (d) and (e), we see saliency map is essential for our transformer-based framework. It can help filter the redundant background noise caused by the attention map.
    Zoom in for best view.}
    \label{Fig.1}
\end{figure}

Inspired by this, we develop a simple yet effective WSSS framework based on Vision Transformer (ViT), termed WegFormer.
Generally, WegFormer has three key parts as follows: an attention map generator based on Deep Taylor Decomposition~(DTD)~\cite{chefer2021transformer}, a soft erasing module, and an efficient potential object mining (EPOM) module.

Among these parts,
(1) the DTD-based attention map generator generates attention maps for the target object. DTD can explain transformer which propagates the relevancy scores through layers~\cite{chefer2021transformer}, which has the high response of the concrete class. We introduce DTD to generate attention maps and integrate them into our WSSS framework.
(2) In the soft erasing module, we introduce a soft rate to narrow the gap between high-response and low-response regions, making the attention map smoother.
(3) EPOM is used to further refine the attention map using saliency maps. Although DTD does an excellent job of distinguishing foreground and background, it also introduces certain noise regions. The saliency maps produced from the offline salient object detector can greatly eliminate these noises, as shown in Figure \ref{Fig.1} (d) and (e).

Equipped with these advanced designs, WegFormer achieves state-of-the-art performance on PASCAL VOC 2012~\cite{everingham2015pascal}.
Notably, WegFormer achieves 66.2\% mIoU on the PASCAL VOC 2012 validation set with VGG-16 as the backbone in the self-training stage, outperforming CNN-based counterparts by more than $0.7\%$ points.
Moreover, when using a heavier backbone ResNet-101 in the self-training stage, WegFormer reaches 70.5\% mIoU, which is the highest of the PASCAL VOC 2012 validation set, significantly outperforming the previous best method NSROM~\cite{yao2021non} by 2.2\% points.

Our contributions can be summarized as follows:

    (1) We propose a simple yet effective Transformer-based WSSS framework, termed WegFormer, which can effectively capture global context information to generate high-quality semantic masks with only image annotation.
    To our knowledge, this is the first work to introduce Transformers into WSSS tasks.
    
    (2) We carefully design three important components in WegFormer, including (1) a DTD-based attention map generator to get an initial attention map for the target object; (2) a soft erasing module to smooth the attention map; (3) an efficient potential object mining (EPOM) to filter background noise in attention map to generate finer pseudo label.
    
    (3) WegFormer achieves state-of-the-art performance on PASCAL VOC 2012 dataset, showing the huge potential of Transformer in WSSS tasks. We hope that WegFormer is a good start for the research in the Transformer-based weakly-supervised segmentation area.
    

\section{Related Work}

\paragraph{Weakly Supervised Semantic Segmentation.}
Weakly supervised semantic segmentation is proposed to learn pixel-level prediction from insufficient labels. Existing solutions are usually based on convolution network~\cite{wei2018revisiting,jiang2019integral}.
There are two main streams. The first stream uses adverse erasing or random erasing in training. \cite{wei2017object} generate CAM in adverse erasing strategy, which finds more extra information not only discriminative regions and introduces a bit of noise. ~\cite{hou2018self} proposed a framework called seeNet, which prevents attention from spreading to the background area. ~\cite{zhang2018adversarial} finds Adversarial Complementary Learning~(ACOL) to localize objects automatically. 
Another stream is spreading CAM from high confidence areas to low. ~\cite{huang2018weakly} train a classification network and then apply region growing algorithm to train segmentation network. ~\cite{fan2020employing} employ multi- estimations to obtain multiple seeds to relieve inaccuracy of a single seed.

Regardless of the validity of the above methods, it is difficult to avoid introducing background noise. 
Recent trends~\cite{kolesnikov2016seed,sun2020mining,yao2021non} attempt to introduce a saliency detector, which is trained offline on other datasets, to generate saliency maps and eliminate background noise.

\paragraph{Transfomers in Computer Vision.}
Transformer has been the dominant architecture in the NLP area and has become popular in the vision community. 
Vision Transformer~(ViT)~\cite{dosovitskiy2020image} is the first work to introduce Transformer into image classification, which divides an image into 16$\times$16 patches. 
IPT~\cite{chen2021pre} is the first transformer pre-train model for low-level vision by combining multi-tasks. 
After that, more and more ViT variants are proposed to extend ViT from different aspects, such as DeiT~\cite{touvron2021training} for efficient training, PVT series~\cite{wang2021pyramid,wang2021pvtv2} and Swin~\cite{liu2021swin} for dense prediction. 
Benefiting from the global receptive field by self-attention~\cite{xie2021segformer}, Transformer is able to capture global context dynamically, which is more friendly for dense prediction tasks such as object detection and semantic segmentation.

Different from previous Transformer frameworks that improve strongly-supervised semantic segmentation~\cite{xie2021segmenting,xie2021segformer}, this paper utilizes Transformer to solve weakly-supervised semantic segmentation with only image-level annotation. 

\begin{figure*}[!t]
    \centering
     \includegraphics[scale=0.7]{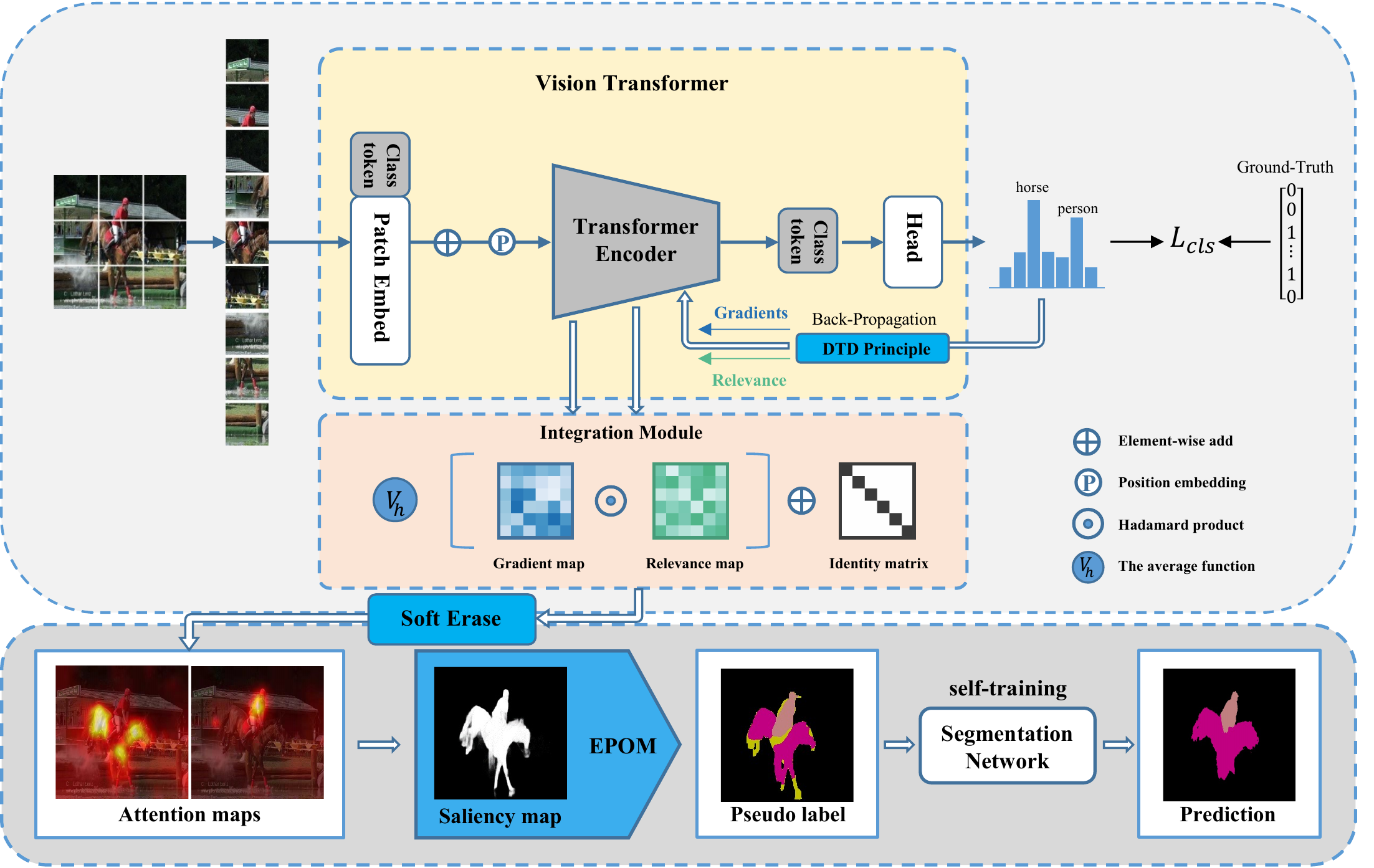}
     \vspace{-1mm}
    \caption{Overview pipeline of WegFormer. 
    First, multi-label classification loss is used to train the vision transformer. 
    Second, the initial relevance score is back-propagation based on the Deep Taylor Decomposition~(DTD) principle to obtain gradient map and relevance map. We integrate gradient maps and relevance maps to get initial attention maps. 
    Third, the soft erase is utilized to smooth the attention map with high and low responses. 
    Forth, efficient potential object mining~(EPOM) avoids pseudo labels from introducing wrong classes and filter out excess background information with saliency maps.
    Finally, the pseudo label is generated and used to self-train the segmentation network.}
    \label{Fig.framework}
    \vspace{-3mm}
\end{figure*}

\paragraph{Neural Network Visualization.}
The deep neural network is a black box, and previous works try to analyze it by visualizing the feature representation in different manners. 
\cite{zhou2016learning} multiplies the weights of the fully connected layer after global average pooling~(GAP) and the spatial feature map before GAP to generate class activation map~(CAM), which can activate the region with objects. 
\cite{selvaraju2017grad} utilizes the gradient of back-propagation to obtain an attention map without modifying the network structure. The CAM and Grad-CAM both work well in the ConvNets but are not suitable for Vision Transformer.
Due to the fundamental architecture difference between Transformer and ConvNet, some attempts have recently tried to visualize Transformers' features.
According to the characteristics of the Transformer, \cite{abnar2020quantifying} proposes two methods to quantify the information flow through self-attention, termed attention rollout and attention flow. The rollout method redistributes all attention scores by considering pairs of attention and assuming a linear combination of attention in a subsequent context. Attention flow is concerned with the maximum flow along the pair-wise attention graph.
\cite{chefer2021transformer} introduces relevance map depending on deep Taylor decomposition and combines the grad of attention map and relevancy map.

\section{Methodology}
The overview framework of WegFormer is illustrated in Figure~\ref{Fig.framework}. 
First, an input RGB image is split into 16$\times$16 patches and fed into a vision transformer classifier. 
Second, Deep Taylor Decomposition~(DTD)~\cite{chefer2021transformer} is used to generate the initial attention maps. 
Third, the soft erase operator smooths attention maps by narrowing the gap between the high-response and the low-response areas. 
Fourth, in EPOM, we use saliency maps to help filter the redundant noise in the background and get refined attention maps and avoid introducing false information to get the final pseudo labels.
Finally, the generated pseudo labels are fed to the segmentation network for self-training to improve the performance further.

\subsection{Attention Map Generation}
Generally, a classification network is trained firstly and we can combine the weights and features to generate class-aware activation maps. After post-processing, the activation map is used as pseudo labels to train a segmentation network to improve the mask quality. 
Different from previous works that use ConvNets as the classifier, we introduce  DeiT~\cite{touvron2021training}, a variant of ViT, as classification networks to capture global contextual information, as illustrated in Figure~\ref{Fig.framework}.

Given images $I \in \mathbb{R}^{C\times H\times W}$ as input, the output of the classification network is a vector $O\in  \mathbb{R}^{1\times c}$, where $c$ indicates the number of the categories.
Due to the fundamental difference between CNN and Transformer, here CAM is not suitable to generate activation maps. Instead, we adopt Deep Taylor Decomposition~(DTD) principle to generate the attention maps~\cite{chefer2021transformer}, which back-propagates the initial relevancy score of the network through all layers and get initial attention map.  

Here we first get the initial relevancy score of per category based on $O$:
\begin{align}
\vspace{-2mm}
    S^{c_i} = \sum(L^{c_i}\odot O),
    \vspace{-2mm}
\end{align}%
where $L^{c_i}\in \mathbb{R}^{1\times c}$ is the one-hot label generated by multi-label ground-truth and $S^{c_i}$ is the initial relevancy scores of the class ${c_i}$. 
Here $\odot$ denotes Hadamard Product.
Following DTD~\cite{chefer2021transformer}, by back-propagating $S^{c_i}$,  we can get relevancy map $R^b \in \mathbb{R}^{h\times(s+1)\times(s+1)}$ for each Transformer block $b$.

Then we dive deep into multi-head self-attention (MHSA) layer~\cite{dosovitskiy2020image}, as shown in Eqn. \ref{eqn:1}:  
\begin{align}
\vspace{-13mm}
  M^{b} = \text{softmax}(\frac{{Q^{b}\cdot K^{b}}^T}{\sqrt{D_\text{head}}}),
\end{align}  
\vspace{-3mm}
\begin{align}
  O^{b} =  M^{b}\cdot V^{b},
\vspace{-3mm}
\label{eqn:1}
\end{align}  
where $Q^{b}$, $K^{b}$, $V^{b}$ means query, key and value in block $b$, and $M^{b}\in \mathbb{R}^{h\times (s+1)\times (s+1)}$, here $s$ is the number of image patch tokens and ``1'' is the class token. $M^{b}$ is self-attention map and $O^{b}$ represents the output of the attention module. $D_\text{head}$ is the head dimension of MHSA.

For $M^{b}$ in each block $b$, we can easily get its gradient $\bigtriangledown M^{b}\in \mathbb{R}^{h\times(s+1)\times(s+1)}$ and relevance $R^{b}\in \mathbb{R}^{h\times(s+1)\times(s+1)}$. 
Then the initial attention map ${A}_{ini}$ can be calculated by:
\begin{align}
\vspace{-8mm}
  {A^{b}}=I + V_h(\bigtriangledown M^{b} \odot R^{b}),
\end{align} 
\vspace{-5mm}
\begin{align}
\label{equ.iniatt}
  A_{ini}=\text{index}(A^{B}, \text{CLS}, \text{axis}=0),
\end{align}
where $I$ is the identity matrix, $V_h(\cdot)$ represents the mean over the ``head'' dimension, and $A^{B} \in \mathbb{R}^{(s+1)\times(s+1)}$. Here $B$ is the last block.

We get the initial attention map $A_{ini} \in \mathbb{R}^{1\times s}$ by indexing (\emph{i.e.}, $\text{index}(\cdot)$) the column corresponding to the class token ``CLS'' according to the horizontal axis.
We reshape and linear interpolation $A_{ini}$ to obtain the final initial attention map  $A_{ini}\in \mathbb{R}^{H\times W}$. 

Unlike the original DTD that utilizes the activation map from all the blocks, here we only take account of the last blocks. We find that the activation map from shallow blocks introduces a certain amount of noise.

\subsection{Soft Erase} \label{sec:se}
Discriminative Region Suppression~(DRS)~\cite{kim2021discriminative} is proposed to suppress discriminative regions and spread to neighboring non-discriminative areas. 
Inspired by DRS, we propose soft erase to narrow the gap between high and low response areas in the initial attention map $A_{ini}$. Unlike DRS that needs to embed into several layers of the network, here soft erase is just a simple post-processing step and does not need to be embedded into the network. 

After get $A_{ini}$ in the above section, we fisrt apply a normalization on $A_{ini}$ and get $\widehat{A} \in {[0, 1]}$.
We then apply soft erase to $\widehat A \in \mathbb{R}^{c\times H \times W} $, which can be written as:
\begin{align}
\vspace{-2mm}
  \widehat{A}_\text{max} = \text{max}(\widehat{A}),\ \ \ \ A_\text{exp} = \text{expand}(\widehat{A}_\text{max}, \widehat{A}),
\end{align}
\vspace{-4mm}
\begin{align}
\label{equ.finalatt}
  A = \text{min}(\widehat{A}, (\widehat A_\text{exp}\odot S_r)),
\end{align}
where $\widehat A_\text{max}\in \mathbb{R}^{c\times 1 \times 1}$, $\widehat A_\text{exp}\in \mathbb{R}^{c\times H \times W}$ and $A\in \mathbb{R}^{c\times H \times W}$. 
$S_r$ is a hyper-parameter and here we set it to $0.55$. 
$\text{max}(\cdot)$, $\text{expand}(\cdot)$, $\text{min}(\cdot)$ mean the maximum, dimension expansion, and minimum function respectively. 
Firstly, a max vector $\widehat A_\text{max}\in \mathbb{R}^{c\times 1 \times 1}$ is chosen from $\widehat{A}$. Then, we expand the dimension of $\widehat A_\text{max}$ to the dimension of $\widehat{A}$. Finally, we compare $\widehat{A}$ and $\widehat A_\text{exp}\odot S_r$ and choose the pixel-wise minimum value to get the attention maps $A$.

\subsection{Efficient Potential Object Mining~(EPOM)}
\cite{yao2021non} proposes POM to use saliency maps for extracting background information. 
Here we find saliency map and attention map generated by DTD are naturally complementary, and saliency maps can largely filter the noise activation in the background.
Note that POM in \cite{yao2021non} uses not only the final heatmap but also utilizes the middle layer heatmaps, which is complicated and low efficient. 
Different from POM, here we propose efficient POM~(EPOM)  that only uses the final attention map, which is more efficient and does not sacrifice performance.

EPOM forces the model to mine potential objects by marking some uncertain pixels in pseudo labels as ``ignored'', which largely avoids introducing wrong labels during self-training.
The detail of EPOM is illustrated as:
\begin{align}
\label{equ.finlabel}
  P_{(x, y)}=\left \{\begin{array}{ll}
  255, & P_{(x, y)}=0~\text{and}~A^{c_i}_{(x,y)}\> > thr^{c_i}\\ P_{(x, y)}, & \text{otherwise} \end{array}\right.
\end{align}
where $P_{(x, y)}$ is the initial pseudo label of pixel position $(x, y)$ generated by comparing the values of $A$, and $c_i$ is current class $i$ of the total number of classes $c$.
In current class $c_i$, if the initial pseudo label $P_{(x, y)}$ equals to $0$ and $A^{c_i}_{(x,y)}$ greater than $thr^{c_i}$, we updated the initial pseudo label $P_{(x, y)}$ as 255 in pixel (x,y). Otherwise, it remains unchanged. Here, ``255'' indicates ``ignored''. 
$thr^{c_i}$ is a threshold for each category that determined dynamically, which can be written as: 
\begin{align}
    thr^{c_i}\!=\!\left \{\begin{array}{ll}
    F_\text{med}(v_{pos}), &\text{if}\ \exists(x,y),\ s.t.\ P_{(x,y)}=c_i \\ F_\text{tq}(v_{pos}), &\text{otherwise} \end{array}\right.
\end{align}
$F_\text{med}(\cdot)$ and $F_\text{tq}(\cdot)$ are functions that obtain the median and top quartile value of position $pos$, respectively. $v_{pos}$ is the value of $pos$. The $pos$ is defined follows: 
\begin{align}
    pos\!=\!\left \{\begin{array}{ll}
    (x,y) | P_{(x,y)}\!=\!c_i, &\text{if}\ \exists(x,y),\ s.t.\ P_{(x,y)}\!=\!c_i \\ (x,y)|A_{(x,y)>T_\text{fg\_thr}}, &\text{otherwise} \end{array}\right.
\end{align}
where $T_\text{fg\_thr}$ is a threshold used to extract foreground information. We choose the position where initial pseudo label equals current class. Otherwise, we choose the position greater than $T_\text{fg\_thr}$ in $A_{(x,y)}$.

\subsection{Self-training with Pseudo Label}
After obtaining the pseudo label, we self-train a segmentation network with the pseudo label to improve the performance. 
Unlike previous work~\cite{yao2021non} that needs to iteratively self-train the segmentation network several times, we only train the segmentation network once.

\section{Experiments}
\subsection{Dataset and Evaluation Metrics}
We use PASCAL VOC 2012~\cite{everingham2015pascal} as the dataset, which is widely used in weakly supervised semantic segmentation. The training set of PASCAL VOC 2012 contains 10582 images with augmentation. Only image-level labels are used during training, and each image contains multiple categories. We report results on validation set~(with 1,449 images) and test set~(with 1,456 images) to compare our approach with other competitive methods.
We use standard mean intersection over union~(mIoU) as the evaluation metric.

\subsection{Implementation Details}
The model is implemented with Pytorch and trained on 1 NVIDIA GeForce RTX 3060 GPU with 12 GB memory. 
In our experiments, we use DeiT-Base~\cite{touvron2021training} as the classification network, which is pre-trained on the ImageNet-1K and we fine-tune it on the PASCAL VOC 2012. During the training phase, we used the AdamW optimizer with a batch size of 16 and the input image is cropped to $224\times224$. During the inference phase, the long side of the input images is cropped to $500~pixels$, and the short side is scaled with an equal proportion to keep the original aspect ratio. The multi-scale inference is also used.

DeepLab-V2/V3/V3+ are used as our segmentation network in the self-training phase. Following common setting~\cite{yao2021non,jiang2019integral}, we compare our results with ResNet-101 and VGG-16 backbone.
 SGD is adopted as the optimizer, and the weight decay is 5e-4. The CRF method is used for post-processing. 

\subsection{Comparisons to the State-of-the-arts}
\begin{table}[h]
    \centering
    \small
    \renewcommand\arraystretch{0.9}
    \scalebox{1.0}{\input{table/sota}}
    \vspace{-7.5mm}
    \caption{Quantitative comparisons to previous state-of-the-art approaches. The left part stands for VGG backbone and the right part stands for ResNet backbone used in segmentation network in self-training stage. $\dagger$ means pre-trained on MS-COCO.}
    \label{Tab.sota}
    \vspace{-5mm}
\end{table}

we compare our methods with existing famous work:  SEC~\cite{kolesnikov2016seed},  STC~\cite{wei2016stc}, ~\cite{roy2017combining}, ~\cite{oh2017exploiting}, AE-PSL~\cite{wei2017object}, WebS-i2~\cite{jin2017webly}, \cite{hong2017weakly}, DCSP~\cite{chaudhry2017discovering}, TPL~\cite{kim2017two}, GAIN~\cite{li2018tell}, DSRG~\cite{huang2018weakly}, MCOF~\cite{wang2018weakly}, AffinityNet~\cite{ahn2018learning}, RDC~\cite{wei2018revisiting}, SeeNet~\cite{hou2018self}, OAA~\cite{jiang2019integral}, ICD ~\cite{fan2020learning}, BES~\cite{chen2020weakly}, ~\cite{fan2020employing}, ~\cite{zhang2020splitting}, MCIS~\cite{sun2020mining}, IRN~\cite{ahn2019weakly}, FickleNet~\cite{lee2019ficklenet}, SSDD~\cite{shimoda2019self}, SEAM~\cite{wang2020self}, SCE~\cite{chang2020weakly}, CONTA~\cite{zhang2020causal}, NSROM~\cite{yao2021non}, DRS\cite{kim2021discriminative}

\begin{figure*}[!t]
    \centering
     \includegraphics[scale=0.6]{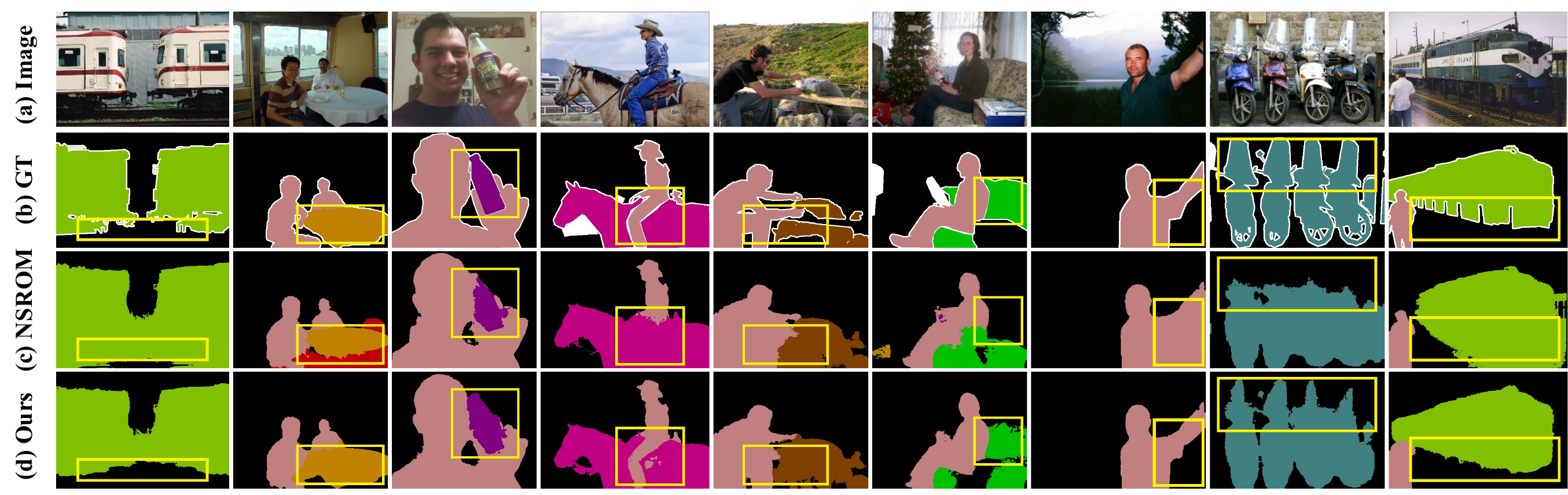}
     \vspace{-2mm}
    \caption{Qualitative segmentation results on the PASCAL VOC 2012 validation set. (a) Input image, (b) Ground truth, (c) Prediction of NSROM~\protect\cite{yao2021non}, (d) Prediction of Ours. We can find that our results are much better than those of NSROM. Best viewed in color.}
    \label{Fig.finalresult}
    \vspace{-2mm}
\end{figure*}

The comparison with state-of-the-art is shown in Table~\ref{Tab.sota}. 
The left column of Table~\ref{Tab.sota} shows the result with the VGG backbone and the right column shows the result with ResNet backbone.
From Table~\ref{Tab.sota}, with the VGG backbone, WegFormer outperforms the NSROM on the test set by $1.2\%$ and the DRS on the validation set by $2.6\%$.
With the ResNet backbone, the upper part terms backbone pre-trained on ImageNet, and WegFormer is $2.2\%$ and $3.7\%$ better than NSROM and DRS on the validation set respectively. 
The bottom part of Table~\ref{Tab.sota}, marked with $\dagger$, means backbone pre-trained on MS-COCO. In the validation set, we are $0.5\%$ ahead of NSROM. 
Overall, for all the backbones, our approach achieves state-of-the-art performance on PASCAL VOC 2012. The qualitative segmentation results on the PASCAL VOC 2012 validation set are shown in Figure~\ref{Fig.finalresult}.

\subsection{Ablation Studies}
In this section we analyze a series of ablation studies and demonstrate the effectiveness of the proposed modules.
\subsubsection{Contribution of Different Components}
As shown in Table~\ref{Tab.sumAblation}, we report the mIoU in validation set with different components. 
Our baseline with initial pseudo label self-training can get $59.5\%$ mIoU. 
By adding soft erase, the resulting increase to $60.0\%$. 
Sal~(DRS) and Sal~(NSROM) means saliency map provided by \cite{kim2021discriminative} and \cite{yao2021non}.
From Table~\ref{Tab.sumAblation} we can find that the saliency map largely boosts up the result to $70.2\%$, with over 10\% improvement. 
EPOM$\ast$ also improves the performance and the result reaches $70.9\%$ finally. 

\begin{table}[!htbp]
\newcommand{\tabincell}[2]{\begin{tabular}{@{}#1@{}}#2\end{tabular}}
\centering
\small
\renewcommand\arraystretch{0.9}
\input{table/sumAblation}
\vspace{-2mm}
\caption{Ablation study of each component in WegFormer. EPOM$\ast$ indicates the method of EPOM without saliency map. }
\label{Tab.sumAblation}
\vspace{-2mm}
\end{table}

\begin{table}[!htbp]
\newcommand{\tabincell}[2]{\begin{tabular}{@{}#1@{}}#2\end{tabular}}
\centering
\small
\renewcommand\arraystretch{0.9}
\input{table/attAblation}
\vspace{-2mm}
\caption{Ablation study of the best blocks to get attention maps. Block indexes are from 0 to 11. Note that 0 means the first block and 11 means the last block.}
\label{Tab.attAblation}
\vspace{-2mm}
\end{table}

\subsubsection{Best Blocks to Get Attention Maps}
Unlike the initial method which integrates all blocks~\cite{chefer2021transformer}, we only adopt the last block as final attention maps as described in equation~\ref{equ.iniatt}. In WegFormer, the total number of blocks in the classifier is 12. 
As shown in Table~\ref{Tab.attAblation}, we find that only taking the last block reaches the best mIoU 59.5\%, which is 1.3\% higher than taking all the blocks.


\begin{figure}[!t]
    \centering
    \scalebox{0.88}{\includegraphics[scale=0.35]{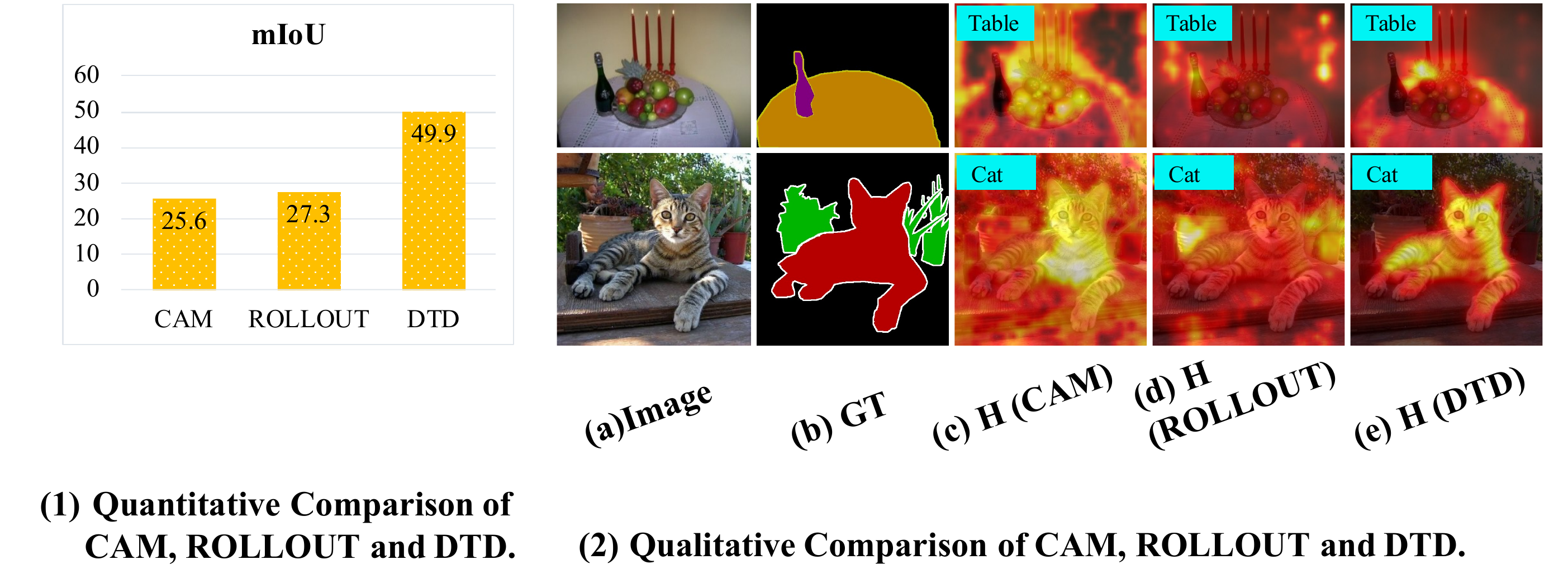}}
     \vspace{-6mm}
    \caption{Comparison between CAM, Rollout, and DTD in Transformer. In sub-figure (1) on the left, the mIoU of DTD is significantly higher than CAM and ROLLOUT, with $+24.3\%$ and $+22.6\%$ respectively.  
    The sub-figure (2) on the right indicates that DTD is significantly better than CAM and ROLLOUT in terms of activating the objects in the feature space. ``H'' means heatmap.
    }
    \label{Fig.CamRolloutDtd}
    \vspace{-1mm}
\end{figure}

\subsubsection{Comparison among CAM, ROLLOUT and DTD}
We compare three methods, CAM, ROLLOUT, and DTD, to generate heatmaps for Transformer in Figure~\ref{Fig.CamRolloutDtd} both quantitatively and qualitatively. 
From Figure~\ref{Fig.CamRolloutDtd}~(1), we see DTD is $24.3\%$ and $22.6\%$ higher than CAM and ROLLOUT respectively. 
Visualization results are shown in Figure~\ref{Fig.CamRolloutDtd}~(2). (c) represents the heatmap obtained by CAM, which has meaningless high responses everywhere in the image. (d) represents the heatmap obtained from ROLLOUT, which cannot distinguish between different classes.
Both CAM and Rollout are not suitable for Transformer. (e) represents the heatmap generated by DTD, which can well excavate the information of the object and capture the contour of the object. 

\begin{figure}[!t]
    \centering
     \includegraphics[scale=0.35]{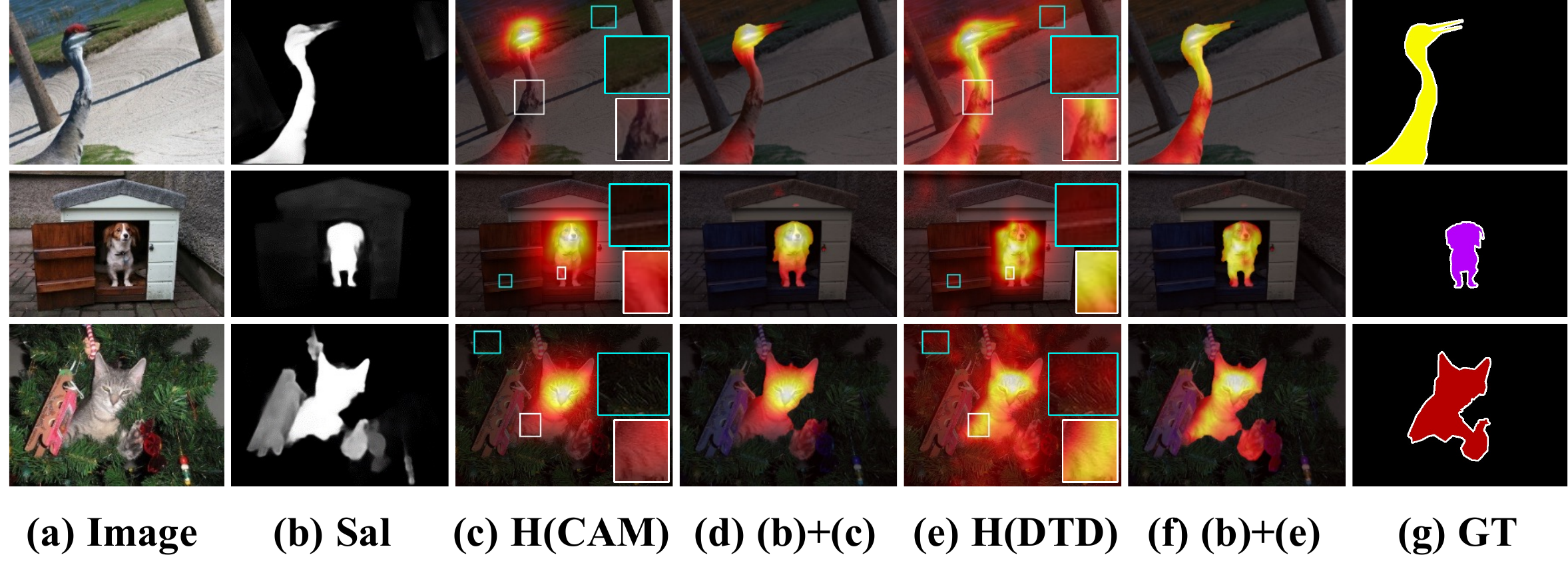}
     \vspace{-3mm}
    \caption{Activation map of Transformer+DTD \textit{vs.} CNN+CAM with or without saliency map. ``H'' means heatmap. As we can see, CNN+CAM focuses on the most discriminative region with less background noise. In contrast, Transformer+DTD can capture the global context information, while also having redundant responses in the background regions. Therefore, combining DTD and saliency map can make more complementary results.}
    \label{Fig.CamDtdSal}
\end{figure}

\begin{table}[!htbp]
\small
\newcommand{\tabincell}[2]{\begin{tabular}{@{}#1@{}}#2\end{tabular}}
\centering
\renewcommand\arraystretch{0.9}
\input{table/seg}
\vspace{-1mm}
\caption{Ablation study of different segmentation framework. Stronger segmentation framework is also important for WSSS.}
\label{Tab.seg}
\end{table}

\subsubsection{Ablation Study of Different Saliency Map}
In Table~\ref{Tab.sumAblation}, we found that the quality of the saliency map also affects the final result.  
Table~2 shows about 1\% gap between saliency map~(DRS) and saliency map~(NSROM), respectively. This inspires us to leverage a higher-quality saliency map to assist weakly-supervised semantic segmentation tasks. We leave it in future research.

\subsubsection{Ablation Study of Stronger Segmentation Networks}
As we can see in Table~\ref{Tab.seg}, more advanced semantic segmentation networks lead to better results. Compared with DeepLab-V2 and DeepLab-V3, DeepLab-V3+ is $1.3\%$ and $0.6\%$ higher on the validation set, respectively. On the test set, we see the similar conclusion.

\subsubsection{Transformer+DTD \textit{vs.} CNN+CAM with Saliency Map}
In Figure~\ref{Fig.CamDtdSal}, we compare the heatmap generated by Transformer+DTD and CNN+CAM with or without saliency maps. Here for Transformer we use Deit-B and for CNN we use ResNet-38.
We find that the CNN+CAM only activates the most discriminative region but fails to capture the whole object. 
In contrast, Transformer+DTD can activate the whole object but also introduces certain background noise simultaneously. 
Therefore, compared with CNN+CAM, saliency maps can better make up for the shortcomings of Transformer+DTD and get high-quality masks.

\section{Conclusion}
In this paper, we propose WegFormer, the first Transformer-based weakly-supervised semantic segmentation framework.
In addition, we introduce three important components: attention map generator based on Deep Taylor Desomposition~(DTD), soft erase module, and efficient potential object mining~(EPOM). These three components could generate high-quality semantic masks as pseudo labels, which significantly boosts the performance.
We hope the proposed WegFormer can serve as a solid baseline to provide a new perspective for weakly supervised semantic segmentation in the Transformer era.

\newpage

\bibliographystyle{named}
\small 
\bibliography{ijcai22}
\end{document}

%% file: table/sota.tex
\begin{tabular}{p{45pt}<{\RaggedRight}p{20pt}<{\centering}p{20pt}<{\centering}p{45pt}<{\RaggedRight}p{20pt}<{\centering}p{20pt}<{\centering}}
\toprule
\multicolumn{3}{c|}{ \rule{0pt}{9pt}\textbf{VGG}}                & \multicolumn{3}{c}{\textbf{ResNet}}              \\ \midrule
 \rule{0pt}{9pt}\textbf{Methods} & \textbf{Val}  & \textbf{Test} & \multicolumn{1}{|l}{\textbf{Methods}} & \textbf{Val}  & \textbf{Test} \\ \midrule
 \rule{0pt}{9pt}SEC              & 50.7          & 51.7          &\multicolumn{1}{|l}{DCSP}             & 60.8          & 61.9          \\
 \rule{0pt}{9pt}STC              & 49.8          & 51.2          & \multicolumn{1}{|l}{DSRG}             & 61.4          & 63.2          \\
 \rule{0pt}{9pt}Roy et al.       & 52.8          & 53.7          & \multicolumn{1}{|l}{MCOF}        & 60.3          & 61.2          \\
 \rule{0pt}{9pt}Oh et al.        & 55.7          & 56.7          & \multicolumn{1}{|l}{AffinityNet}      & 61.7          & 63.7          \\
 \rule{0pt}{9pt}AE-PSL           & 55.0          & 55.7          & \multicolumn{1}{|l}{SeeNet}           & 63.1          & 62.8          \\
 \rule{0pt}{9pt}WebS-i2          & 53.4          & 55.3          & \multicolumn{1}{|l}{IRN}              & 63.5          & 64.8          \\
 \rule{0pt}{9pt}Hong et al.      & 58.1          & 58.7          & \multicolumn{1}{|l}{FickleNet}        & 64.9          & 65.3          \\
 \rule{0pt}{9pt}DCSP             & 58.6          & 59.2          & \multicolumn{1}{|l}{OAA}              & 65.2          & 66.4          \\
 \rule{0pt}{9pt}TPL              & 53.1          & 53.8          & \multicolumn{1}{|l}{SSDD}             & 64.9          & 65.5          \\
 \rule{0pt}{9pt}GAIN             & 55.3          & 56.8          & \multicolumn{1}{|l}{SEAM}             & 64.5          & 65.7          \\
 \rule{0pt}{9pt}DSRG             & 59.0          & 60.4          & \multicolumn{1}{|l}{SCE}              & 66.1          & 65.9          \\
 \rule{0pt}{9pt}MCOF             & 56.2          & 57.6          & \multicolumn{1}{|l}{ICD}              & 67.8          & 68.0          \\
 \rule{0pt}{9pt}AffinityNet      & 58.4          & 60.5          & \multicolumn{1}{|l}{Zhang et al.}     & 66.6          & 66.7          \\
 \rule{0pt}{9pt}RDC              & 60.4          & 60.8          & \multicolumn{1}{|l}{Fan et al.}       & 67.2          & 66.7          \\
 \rule{0pt}{9pt}SeeNet           & 63.1          & 62.8          & \multicolumn{1}{|l}{MCIS}             & 66.2          & 66.9          \\
 \rule{0pt}{9pt}OAA              & 63.1          & 62.8          & \multicolumn{1}{|l}{BES}              & 65.7          & 66.6          \\
 \rule{0pt}{9pt}ICD              & 64.0          & 63.9          & \multicolumn{1}{|l}{CONTA}            & 66.1          & 66.7          \\
 \rule{0pt}{9pt}BES              & 60.1          & 61.1          & \multicolumn{1}{|l}{DRS}              & 66.8          & 67.4          \\
 \rule{0pt}{9pt}Fan et al.       & 64.6          & 64.2          & \multicolumn{1}{|l}{NSROM}            & 68.3          & 68.5          \\
 \rule{0pt}{9pt}Zhang et al.     & 63.7          & 64.5          & \multicolumn{1}{|l}{\textbf{Ours}}    & \textbf{70.5} & \textbf{70.3} \\ \cline{4-6} 
 \rule{0pt}{9pt}MCIS             & 63.5          & 63.6          & \multicolumn{1}{|l}{OAA$\dagger$}             & 67.4          & —             \\
 \rule{0pt}{9pt}DRS              & 63.6          & 64.4          & \multicolumn{1}{|l}{\textbf{DRS$\dagger$}}    & \textbf{71.2} & \textbf{71.4} \\
 \rule{0pt}{9pt}NSROM            & 65.5          & 65.3          & \multicolumn{1}{|l}{NSROM$\dagger$}           & 70.4          & 70.2          \\
 \rule{0pt}{9pt}\textbf{Ours}    & \textbf{66.2} & \textbf{66.5} & \multicolumn{1}{|l}{Ours$\dagger$}            & 70.9          & 70.5          \\ \bottomrule
\label{Tab.quan}
\end{tabular}

%% file: table/sumAblation.tex
\begin{tabular}{p{42pt}<{\centering}p{35pt}<{\centering}p{46pt}<{\centering}p{30pt}<{\centering}p{18pt}<{\centering}}
\toprule
 \rule{0pt}{8pt} \textbf{Soft Erase}&\textbf{Sal~(DRS)}&\textbf{Sal~(NSROM)}&\textbf{EPOM$\ast$}&\textbf{mIoU}\\
\midrule
 \rule{0pt}{8pt} \text{-} & \text{-} & \text{-} & \text{-} &  \text{59.5}\\
 \rule{0pt}{8pt} \checkmark & \text{-} & \text{-} & \text{-} &  \text{60.0}\\
 \rule{0pt}{8pt} \checkmark & \checkmark & \text{-} & \text{-} & \text{70.2}\\
 \rule{0pt}{8pt} \checkmark & \checkmark & \text{-} & \checkmark & \text{70.9}\\
 \midrule
 \rule{0pt}{8pt} \checkmark & \text{-} & \checkmark & \text{-} & \text{68.9}\\
 \rule{0pt}{8pt} \checkmark & \text{-} & \checkmark & \checkmark & \text{69.8}\\

\bottomrule
\end{tabular}

%% file: table/attAblation.tex
\begin{tabular}{p{130pt}<{\RaggedRight}p{40pt}<{\centering}}
\toprule
 \rule{0pt}{8pt}\textbf{Blocks} & \textbf{mIoU}\\
\midrule
 \rule{0pt}{8pt}\text{0, 1, 2, 3, 4, 5, 6, 7, 8, 9, 10, 11} & \text{58.2}\\
\midrule
 \rule{0pt}{8pt}\text{3,4,5,6,7,8,9,10,11} & \text{58.4}\\
\midrule
 \rule{0pt}{8pt}\text{5,6,7,8,9,10,11} & \text{58.2}\\
\midrule
 \rule{0pt}{8pt}\text{7,8,9,10,11} & \text{58.8}\\
\midrule
 \rule{0pt}{8pt}\text{9,10,11} & \text{59.0}\\
\midrule
 \rule{0pt}{8pt}\text{11} & \text{59.5}\\
\bottomrule
\end{tabular}

%% file: table/seg.tex
\begin{tabular}{p{60pt}<{\raggedright}p{55pt}<{\centering}p{31pt}<{\centering}}

\toprule
 \rule{0pt}{9pt}   \textbf{Network}   & \textbf{val} & \textbf{test}\\
\midrule
 \rule{0pt}{9pt}   \text{DeepLab-V2}  & \text{69.2}  & \text{69.7}\\
\midrule
 \rule{0pt}{9pt}   \text{DeepLab-V3} & \text{69.3}    & \text{69.9}\\
\midrule
 \rule{0pt}{9pt}   \text{DeepLab-V3+} & \text{70.5}  & \text{70.3}\\
\bottomrule
\end{tabular}